\documentclass[conference]{IEEEtran}
\IEEEoverridecommandlockouts
\usepackage{cite}
\usepackage{amsmath,amssymb,amsfonts}
\usepackage{algorithmic}
\usepackage{graphicx}
\usepackage{textcomp}
\usepackage{xcolor}
\usepackage{siunitx}
\usepackage{svg}
\usepackage{acronym}

\acrodef{AC}[AC]{Coefficient of Variation}
\acrodef{ADC}[ADC]{Analog-to-Digital Converter}
\acrodef{ADEXP}[AdExp-I\&F]{Adaptive-Exponential Integrate and Fire}
\acrodef{ADM}[ADM]{Asynchronous Delta Modulator}
\acrodef{AER}[AER]{Address-Event Representation}
\acrodef{AEX}[AEX]{AER EXtension board}
\acrodef{AE}[AE]{Address-Event}
\acrodef{AFM}[AFM]{Atomic Force Microscope}
\acrodef{AGC}[AGC]{Automatic Gain Control}
\acrodef{AI}[AI]{Artificial Intelligence}
\acrodef{AMDA}[AMDA]{AER Motherboard with D/A converters}
\acrodef{AMPA}[AMPA]{$\alpha$-amino-3-hydroxy-5-methyl-4-isoxazolepropionic acid}
\acrodef{ANN}[ANN]{Artificial Neural Network}
\acrodef{API}[API]{Application Programming Interface}
\acrodef{APMOM}[APMOM]{Alternate Polarity Metal On Metal}
\acrodef{ARM}[ARM]{Advanced RISC Machine}
\acrodef{ASIC}[ASIC]{Application Specific Integrated Circuit}
\acrodef{AdExp}[AdExp-IF]{Adaptive Exponential Integrate-and-Fire}
\acrodef{BCM}[BMC]{Bienenstock-Cooper-Munro}
\acrodef{BD}[BD]{Bundled Data}
\acrodef{BEOL}[BEOL]{Back-end of Line}
\acrodef{BG}[BG]{Bias Generator}
\acrodef{BMI}[BMI]{Brain-Machince Interface}
\acrodef{BPTT}[BPTT]{Backpropagation through time}
\acrodef{BTB}[BTB]{band-to-band tunnelling}
\acrodef{CAD}[CAD]{Computer Aided Design}
\acrodef{CAM}[CAM]{Content Addressable Memory}
\acrodef{CAN}[CAN]{Continuous attractor network}
\acrodef{CAVIAR}[CAVIAR]{Convolution AER Vision Architecture for Real-Time}
\acrodef{CA}[CA]{Cortical Automaton}
\acrodef{CCN}[CCN]{Cooperative and Competitive Network}
\acrodef{CDR}[CDR]{Clock-Data Recovery}
\acrodef{CFC}[CFC]{Current to Frequency Converter}
\acrodef{CHP}[CHP]{Communicating Hardware Processes}
\acrodef{CMIM}[CMIM]{Metal-insulator-metal Capacitor}
\acrodef{CML}[CML]{Current Mode Logic}
\acrodef{CMOL}[CMOL]{Hybrid CMOS nanoelectronic circuits}
\acrodef{CMOS}[CMOS]{complementary metal–oxide–semiconductor}
\acrodef{CNN}[CNN]{Convolutional Neural Network}
\acrodef{CNS}[CNS]{central nervous system}
\acrodef{COTS}[COTS]{Commercial Off-The-Shelf}
\acrodef{CPG}[CPG]{Central Pattern Generator}
\acrodef{CPLD}[CPLD]{Complex Programmable Logic Device}
\acrodef{CPU}[CPU]{Central Processing Unit}
\acrodef{CSM}[CSM]{Cortical State Machine}
\acrodef{CSP}[CSP]{Constraint Satisfaction Problem}
\acrodef{CTXCTL}[CTXCTL]{CortexControl}
\acrodef{CV}[CV]{Coefficient of Variation}
\acrodef{DAC}[DAC]{Digital to Analog Converter}
\acrodef{DAS}[DAS]{Dynamic Auditory Sensor}
\acrodef{DAVIS}[DAVIS]{Dynamic and Active Pixel Vision Sensor}
\acrodef{DBN}[DBN]{Deep Belief Network}
\acrodef{DBS}[DBS]{Deep Brain Stimulation}
\acrodef{DFA}[DFA]{Deterministic Finite Automaton}
\acrodef{DIBL}[DIBL]{drain-induced-barrier-lowering}
\acrodef{DI}[DI]{delay insensitive}
\acrodef{DMA}[DMA]{Direct Memory Access}
\acrodef{DNF}[DNF]{Dynamic Neural Field}
\acrodef{DNN}[DNN]{Deep Neural Network}
\acrodef{DOF}[DOF]{Degrees of Freedom}
\acrodef{DPE}[DPE]{Dynamic Parameter Estimation}
\acrodef{DPI}[DPI]{Differential Pair Integrator}
\acrodef{DRAM}[DRAM]{Dynamic Random Access Memory}
\acrodef{DRRZ}[DR-RZ]{Dual-Rail Return-to-Zero}
\acrodef{DR}[DR]{Dual Rail}
\acrodef{DSP}[DSP]{Digital Signal Processor}
\acrodef{DVS}[DVS]{Dynamic Vision Sensor}
\acrodef{DYNAP}[DYNAP]{Dynamic Neuromorphic Asynchronous Processor}
\acrodef{EBL}[EBL]{Electron Beam Lithography}
\acrodef{ECG}[ECG]{Electrocardiography}
\acrodef{ECoG}[ECoG]{Electrocorticography}
\acrodef{EDVAC}[EDVAC]{Electronic Discrete Variable Automatic Computer}
\acrodef{EEG}[EEG]{Electroencephalography}
\acrodef{EIN}[EIN]{Excitatory-Inhibitory Network}
\acrodef{EMG}[EMG]{Electromyography}
\acrodef{EM}[EM]{Expectation Maximization}
\acrodef{EOG}[EOG]{Electrooculogram}
\acrodef{EPSC}[EPSC]{Excitatory Post-Synaptic Current}
\acrodef{EPSP}[EPSP]{Excitatory Post-Synaptic Potential}
\acrodef{EZ}[EZ]{Epileptogenic Zone}
\acrodef{FDSOI}[FDSOI]{Fully-Depleted Silicon on Insulator}
\acrodef{FET}[FET]{Field-Effect Transistor}
\acrodef{FFT}[FFT]{Fast Fourier Transform}
\acrodef{FI}[F-I]{Frequency-Current}
\acrodef{FMA}[FMA]{Floating microelectrode array} 
\acrodef{FNN}[FNN]{Feed-forward Neural Network}
\acrodef{FPGA}[FPGA]{Field Programmable Gate Array}
\acrodef{FR}[FR]{Fast Ripple}
\acrodef{FSA}[FSA]{Finite State Automaton}
\acrodef{FSM}[FSM]{Finite State Machine}
\acrodef{GABA}[GABA]{$\gamma$-aminobutanoic acid}
\acrodef{GIDL}[GIDL]{gate-induced-drain-leakage}
\acrodef{GOPS}[GOPS]{Giga-Operations per Second}
\acrodef{GPU}[GPU]{Graphical Processing Unit}
\acrodef{GT}[GT]{Ground Truth}
\acrodef{GUI}[GUI]{Graphical User Interface}
\acrodef{HAL}[HAL]{Hardware Abstraction Layer}
\acrodef{HFO}[HFO]{High Frequency Oscillation}
\acrodef{HH}[H\&H]{Hodgkin \& Huxley}
\acrodef{HMM}[HMM]{Hidden Markov Model}
\acrodef{HRS}[HRS]{High-Resistive State}
\acrodef{HR}[HR]{Human Readable}
\acrodef{HSE}[HSE]{Handshaking Expansion}
\acrodef{HW}[HW]{Hardware}
\acrodef{ICT}[ICT]{Information and Communication Technology}
\acrodef{IC}[IC]{Integrated Circuit}
\acrodef{IF2DWTA}[IF2DWTA]{Integrate \& Fire 2--Dimensional WTA}
\acrodef{IFSLWTA}[IFSLWTA]{Integrate \& Fire Stop Learning WTA}
\acrodef{IF}[I\&F]{Integrate-and-Fire}
\acrodef{IMU}[IMU]{Inertial Measurement Unit}
\acrodef{INCF}[INCF]{International Neuroinformatics Coordinating Facility}
\acrodef{INI}[INI]{Institute of Neuroinformatics}
\acrodef{IO}[I/O]{Input/Output}
\acrodef{IPSC}[IPSC]{Inhibitory Post-Synaptic Current}
\acrodef{IPSP}[IPSP]{Inhibitory Post-Synaptic Potential}
\acrodef{IP}[IP]{Intellectual Property}
\acrodef{ISI}[ISI]{Inter-Spike Interval}
\acrodef{IoT}[IoT]{Internet of Things}
\acrodef{ITL}[ITL]{In-The-Loop}
\acrodef{JFLAP}[JFLAP]{Java - Formal Languages and Automata Package}
\acrodef{LEDR}[LEDR]{Level-Encoded Dual-Rail}
\acrodef{LFP}[LFP]{Local Field Potential}
\acrodef{LIFE}[LIFE]{Longitudinal Intrafascicular Electrodes}
\acrodef{LIF}[LIF]{Leak Integrate-and-Fire}
\acrodef{LLC}[LLC]{Low Leakage Cell}
\acrodef{LNA}[LNA]{Low-Noise Amplifier}
\acrodef{LPF}[LPF]{Low Pass Filter}
\acrodef{LRS}[LRS]{Low-Resistive State}
\acrodef{LSM}[LSM]{Liquid State Machine}
\acrodef{LTD}[LTD]{Long Term Depression}
\acrodef{LTI}[LTI]{Linear Time-Invariant}
\acrodef{LTP}[LTP]{Long Term Potentiation}
\acrodef{LTU}[LTU]{Linear Threshold Unit}
\acrodef{LUT}[LUT]{Look-Up Table}
\acrodef{LVDS}[LVDS]{Low Voltage Differential Signaling}
\acrodef{MCMC}[MCMC]{Markov-Chain Monte Carlo}
\acrodef{MEA}[MEA]{Multielectrode Arrays}
\acrodef{MEMS}[MEMS]{Micro Electro Mechanical System}
\acrodef{MFR}[MFR]{Mean Firing Rate}
\acrodef{MIM}[MIM]{Metal Insulator Metal}
\acrodef{MLP}[MLP]{Multilayer Perceptron}
\acrodef{ML}[ML]{Machine Learning}
\acrodef{MOSCAP}[MOSCAP]{Metal Oxide Semiconductor Capacitor}
\acrodef{MOSFET}[MOSFET]{Metal Oxide Semiconductor Field-Effect Transistor}
\acrodef{MOS}[MOS]{Metal Oxide Semiconductor}
\acrodef{MRI}[MRI]{Magnetic Resonance Imaging}
\acrodef{NCS}[NCS]{Neuromorphic Cognitive Systems}
\acrodef{NDFSM}[NDFSM]{Non-deterministic Finite State Machine} 
\acrodef{ND}[ND]{Noise-Driven}
\acrodef{NEF}[NEF]{Neural Engineering Framework}
\acrodef{NHML}[NHML]{Neuromorphic Hardware Mark-up Language}
\acrodef{NIL}[NIL]{Nano-Imprint Lithography}
\acrodef{NI}[NI]{Neural Interface}
\acrodef{NMDA}[NMDA]{N-Methyl-D-Aspartate}
\acrodef{NME}[NE]{Neuromorphic Engineering}
\acrodef{NN}[NN]{Neural Network}
\acrodef{NOC}[NoC]{Network-on-Chip}
\acrodef{NRZ}[NRZ]{Non-Return-to-Zero}
\acrodef{NSM}[NSM]{Neural State Machine}
\acrodef{OR}[OR]{Operating Room}
\acrodef{OTA}[OTA]{Operational Transconductance Amplifier}
\acrodef{PCB}[PCB]{Printed Circuit Board}
\acrodef{PCHB}[PCHB]{Pre-Charge Half-Buffer}
\acrodef{PCM}[PCM]{Phase Change Memory}
\acrodef{PCA}[PCA]{Personal Component Analysis}

\acrodef{PC}[PC]{Personal Computer}
\acrodef{PE}[PE]{Phase Encoding}
\acrodef{PFA}[PFA]{Probabilistic Finite Automaton}
\acrodef{PFC}[PFC]{prefrontal cortex}
\acrodef{PFM}[PFM]{Pulse Frequency Modulation}
\acrodef{PNI}[PNI]{peripheral nerve interface}
\acrodef{PNS}[PNS]{peripheral nervous system}
\acrodef{PPG}[PPG]{Photoplethysmography}
\acrodef{PR}[PR]{Production Rule}
\acrodef{PSC}[PSC]{Post-Synaptic Current}
\acrodef{PSP}[PSP]{Post-Synaptic Potential}
\acrodef{PSTH}[PSTH]{Peri-Stimulus Time Histogram}
\acrodef{PV}[PV]{Parvalbumin}
\acrodef{QDI}[QDI]{Quasi Delay Insensitive}
\acrodef{RAM}[RAM]{Random Access Memory}
\acrodef{RA}[RA]{Resected Area}
\acrodef{RDF}[RDF]{random dopant fluctuation}
\acrodef{RELU}[ReLu]{Rectified Linear Unit}
\acrodef{RLS}[RLS]{Recursive Least-Squares}
\acrodef{RMSE}[RMSE]{Root Mean Squared-Error}
\acrodef{RMS}[RMS]{Root Mean Squared}
\acrodef{RNN}[RNN]{Recurrent Neural Networks}
\acrodef{RNN}[RNN]{Recurrent Neural Network}
\acrodef{ROLLS}[ROLLS]{Reconfigurable On-Line Learning Spiking}
\acrodef{RRAM}[R-RAM]{Resistive Random Access Memory}
\acrodef{RSA}[RSA]{Respiratory Sinus Arrhythmia}
\acrodef{R}[R]{Ripples}
\acrodef{SAC}[SAC]{Selective Attention Chip}
\acrodef{SAT}[SAT]{Boolean Satisfiability Problem}
\acrodef{SCI}[SCI]{Spinal Cord Injury}
\acrodef{SCX}[SCX]{Silicon CorteX}
\acrodef{SD}[SD]{Signal-Driven}
\acrodef{SEM}[SEM]{Spike-based Expectation Maximization}
\acrodef{SLAM}[SLAM]{Simultaneous Localization and Mapping}
\acrodef{SNN}[SNN]{Spiking Neural Network}
\acrodef{SNR}[SNR]{Signal to Noise Ratio}
\acrodef{SOC}[SOC]{System-On-Chip}
\acrodef{SOI}[SOI]{Silicon on Insulator}
\acrodef{SOZ}[SOZ]{Seizure Onset Zone}
\acrodef{SP}[SP]{Separation Property}
\acrodef{SRAM}[SRAM]{Static Random Access Memory}
\acrodef{PYR}[PYR]{Pyramidal}
\acrodef{SST}[SST]{Somatostatin}

\acrodef{STDP}[STDP]{Spike-Timing Dependent Plasticity}
\acrodef{STD}[STD]{Short-Term Depression}
\acrodef{STP}[STP]{Short-Term Plasticity}
\acrodef{STT-MRAM}[STT-MRAM]{Spin-Transfer Torque Magnetic Random Access Memory}
\acrodef{STT}[STT]{Spin-Transfer Torque}
\acrodef{SVM}[SVM]{Support Vector Machine}
\acrodef{SW}[SW]{Software}
\acrodef{TCAM}[TCAM]{Ternary Content-Addressable Memory}
\acrodef{TFT}[TFT]{Thin Film Transistor}
\acrodef{TIME}[TIME]{Transverse Intrafascicular Multichannel Electrode}
\acrodef{TLE}[TLE]{Temporal Lobe Epilepsy}
\acrodef{UEA}[UEA]{Utah electrode array}
\acrodef{USB}[USB]{Universal Serial Bus}
\acrodef{USEA}[USEA]{Utah Slanted Electrode Array}
\acrodef{VHDL}[VHDL]{VHSIC Hardware Description Language}
\acrodef{VIP}[VIP]{Vasoactive Intestinal Peptide}
\acrodef{VLSI}[VLSI]{Very Large Scale Integration}
\acrodef{VNS}[VNS]{Vagal Nerve Stimulation}
\acrodef{VOR}[VOR]{Vestibulo-Ocular Reflex}
\acrodef{WCST}[WCST]{Wisconsin Card Sorting Test}
\acrodef{WTA}[WTA]{Winner-Take-All}
\acrodef{XML}[XML]{eXtensible Mark-up Language}
\acrodef{divmod3}[DIVMOD3]{divisibility of a number by three}
\acrodef{hWTA}[hWTA]{hard Winner-Take-All}
\acrodef{iEEG}[iEEG]{intracranial electroencephalography}
\acrodef{sWTA}[sWTA]{soft Winner-Take-All}

\acrodef{PVA}[PVA]{Population Vector Average}

\def\BibTeX{{\rm B\kern-.05em{\sc i\kern-.025em b}\kern-.08em
    T\kern-.1667em\lower.7ex\hbox{E}\kern-.125emX}}
\begin{document}

\title{Neuromorphic Spiking  Ring Attractor for Proprioceptive Joint-State Estimation}

\author{
 \IEEEauthorblockN{Federica Ferrari$^{1*}$, Flavia Davidhi$^{2,3*}$, Bernard Maacaron$^{1*}$, Alberto Motta$^{4}$, Luuk van Keeken$^{5}$,\\ Elisa Donati$^{3}$, Giacomo Indiveri$^{3}$, Chiara De Luca$^{3,6}$, Chiara Bartolozzi$^{1}$}
 \IEEEauthorblockA{
 $^{1}$Istituto Italiano di Tecnologia, Genoa, Italy, $^{2}$Department of Neurosurgery, University Hospital Zürich, Switzerland\\
$^{3}$Institute of Neuroinformatics, Univ. of Zurich and ETH Zurich, Switzerland\\
$^{4}$DIEEI, University of Catania
Catania, Italy,
$^{5}$IDLAB, University of Antwerp, and AI \& Algorithms, imec, Leuven, Belgium\\
$^{6}$Digital Society Initiative, University of Zurich, Switzerland\\
* these authors contributed equally}
}

\maketitle
\begin{abstract}

Maintaining stable internal representations of continuous variables is fundamental for effective robotic control. Continuous attractor networks provide a biologically inspired mechanism for encoding such variables, yet neuromorphic realizations have rarely addressed proprioceptive estimation under resource constraints. This work introduces a spiking ring-attractor network representing a robot joint angle through self-sustaining population activity. Local excitation and broad inhibition support a stable activity bump, while velocity-modulated asymmetries drive its translation and boundary conditions confine motion within mechanical limits. The network reproduces smooth trajectory tracking and remains stable near joint limits, showing reduced drift and improved accuracy compared to unbounded models. Such compact hardware-compatible implementation preserves multi-second stability demonstrating a near-linear relationship between bump velocity and synaptic modulation.
\end{abstract}


\section{Introduction}
Accurate models of body state are essential for autonomous robots to plan and control movements effectively. In particular, estimating head orientation relative to the world (head pose) enables vision stabilization, gaze coordinating and multisensory integration during motion~\cite{falotico2017head}. In biological systems, these functions are supported by head-direction circuits that maintain a persistent representation of orientation through recurrent neural dynamics~\cite{stone2017anatomically, seelig2015neural}.
\acp{CAN} provide a canonical mechanism for representing continuous variables through localized, self-sustaining activity bumps~\cite{Song_Wang_2005,Seeholzer_Deger_Gerstner_2019}. A ring attractor is a specific type of \ac{CAN} in which the activity bump moves along a circular manifold, naturally encoding angular variables such as head orientation.
Experimental evidence for ring-like attractor dynamics has been observed in the fly central complex and mammalian head-direction systems~\cite{Kim_Rouault_Druckmann_Jayaraman_2017}, where recurrent excitation and broad inhibition stabilize these activity patterns over time.
In contrast, conventional robotic state estimation typically relies on sensor fusion algorithms that, while accurate, are computationally costly and power demanding~\cite{olofsson2015sensor}. Neuromorphic hardware offers a biologically-inspired alternative, enabling asynchronous, event-driven computation with low energy cost and adaptive dynamics~\cite{bartolozzi2022embodied, chicca2014neuromorphic}, and thus represents a promising substrate for real-time robotic state estimation.

Neuromorphic implementations of attractor networks have been explored across a variety of platforms and applied to domains such as navigation, orientation and control.
Several of these works target hardware realizations on digital neuromorphic systems such as Loihi~\cite{kreiser2020chip,Tang_Shah_Michmizos_2019}, typically relying on large neuron populations or multilayer architectures to ensure stability and noise robustness~\cite{ Moradi_Qiao_Stefanini_Indiveri_2018,fisher2019sensorimotor}. While effective, these approaches incur significant computational and resource overhead, are often tailored to sensory or motor processing rather than proprioceptive estimation~\cite{Zhao_Risi_Monforte_Bartolozzi_Indiveri_Donati_2020,Richter_Wu_Whatley_Köstinger_Nielsen_Qiao_Indiveri_2024}, and frequently address velocity encoding through additional network layers. Moreover, many existing models assume ideal circular manifolds without accounting for mechanical joint limits~\cite{Samsonovich_McNaughton_1997,noormanMaintainingUpdatingAccurate2024} or lack an explicit calibration between physical velocity inputs and attractor dynamics~\cite{kreiser2020chip}.
These factors leave open the question whether a compact, bounded, spiking attractor can achieve accurate proprioceptive estimation while remaining compatible the limited resources of with mixed-signal neuromorphic hardware.
In this work, we address this gap proposing a bounded spiking ring-attractor for proprioceptive pose estimation based on synaptic modulation. Our design combines a biologically inspired architecture (i.e. local recurrent excitation with broad inhibition) with a tunable mapping between physical velocity and bump dynamics. To evaluate hardware compatibility, we implement  a compact single-layer network (in contrast to large-scale multilayer attractors~\cite{kreiser2020chip,fisher2019sensorimotor}) on the mixed-signal DYNAP-SE1 processor in a computer-in-the-loop setup. Both simulation and on-chip tests demonstrate that the network maintains a stable joint-state representation and preserves accuracy across velocity regimes. Overall, we contribute a biologically grounded algorithm for proprioceptive estimation, a calibrated velocity-to-bump relation, and a validation on mixed-signal neuromorphic hardware under tight resource constraints.
\begin{figure*}
    \centering
    \includegraphics[width=0.9\textwidth]{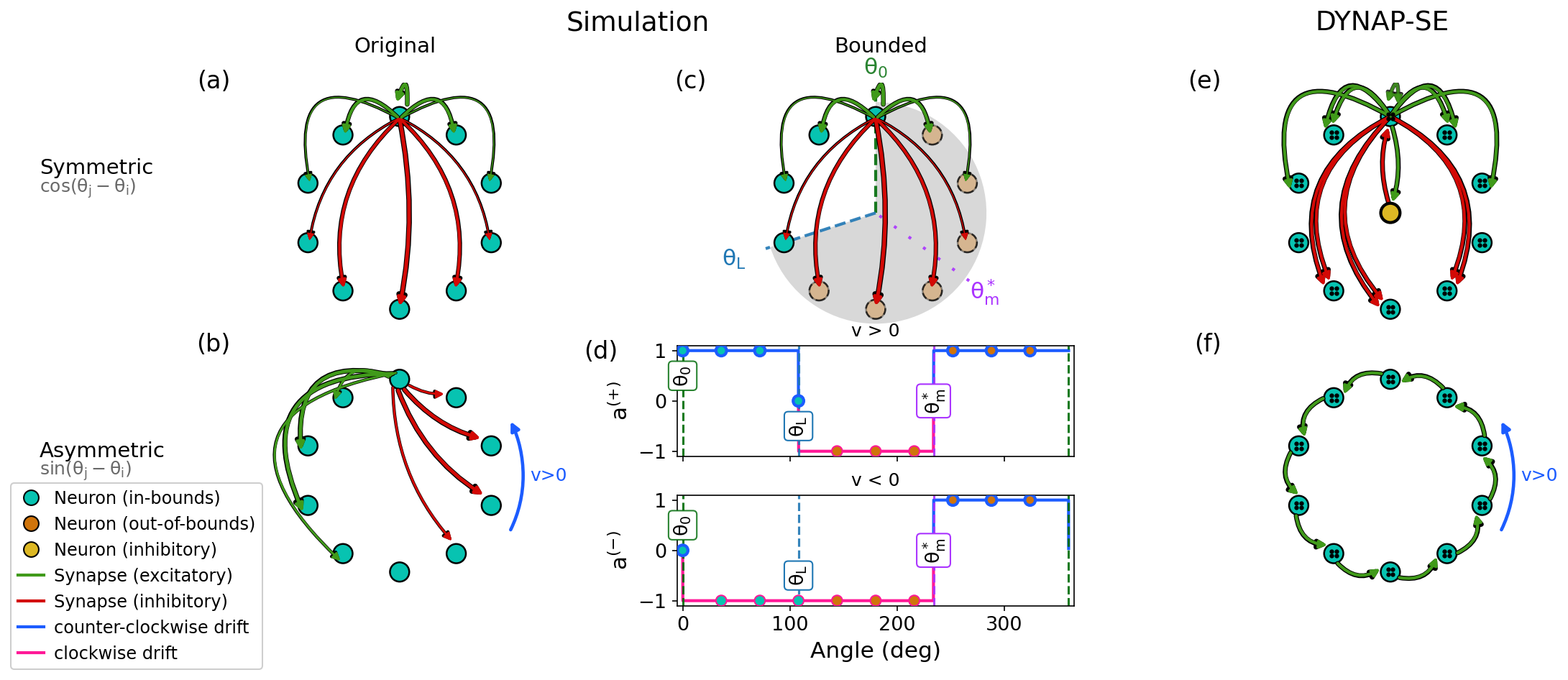}
    \caption{\textbf{Ring attractor architectures used in the simulation (with and without boundaries) and on DYNAP-SE.} Top row: symmetric profiles across architectures, bottom row: asymmetric profiles. Connections are only shown for neuron on top. (a) The symmetric cosine-shaped profile in the original model. (b) The asymmetric sinusoidal profile in the original model. (c) Symmetric cosine-shaped profile with boundaries (\(\theta_0\),\(\theta_l\)) and the middle out-of-bound angle \(\theta_m^*\); gray area indicates the out-of-bound region.(d) The asymmetric profile in the bounded model obtained modulating the original profile through \(a^+\) and \(a^-\). (e) DYNAP-SE symmetric profile, where number of connections increases with connection strength. (f) asymmetric profile in the DYNAP-SE implementation, where number of connections increase with the velocity intensity.}\label{fig:model_schematic}
\end{figure*}

\section{Materials \& Methods}
The model extends the ring attractor framework~\cite{noormanMaintainingUpdatingAccurate2024} to a spiking implementation with equivalent dynamics. Local excitation and global inhibition sustain a stable activity bump, whose angular estimate is read out via \ac{PVA}~\cite{georgopoulosNeuronalPopulationCoding1986}. Network gains were first tuned by grid search and then refined through gradient-based calibration to map simulated joint velocity to the bump’s angular velocity on the neural ring.
\begin{figure*}[ht]
\centering
\includegraphics[width=1\textwidth,height=5.3cm]{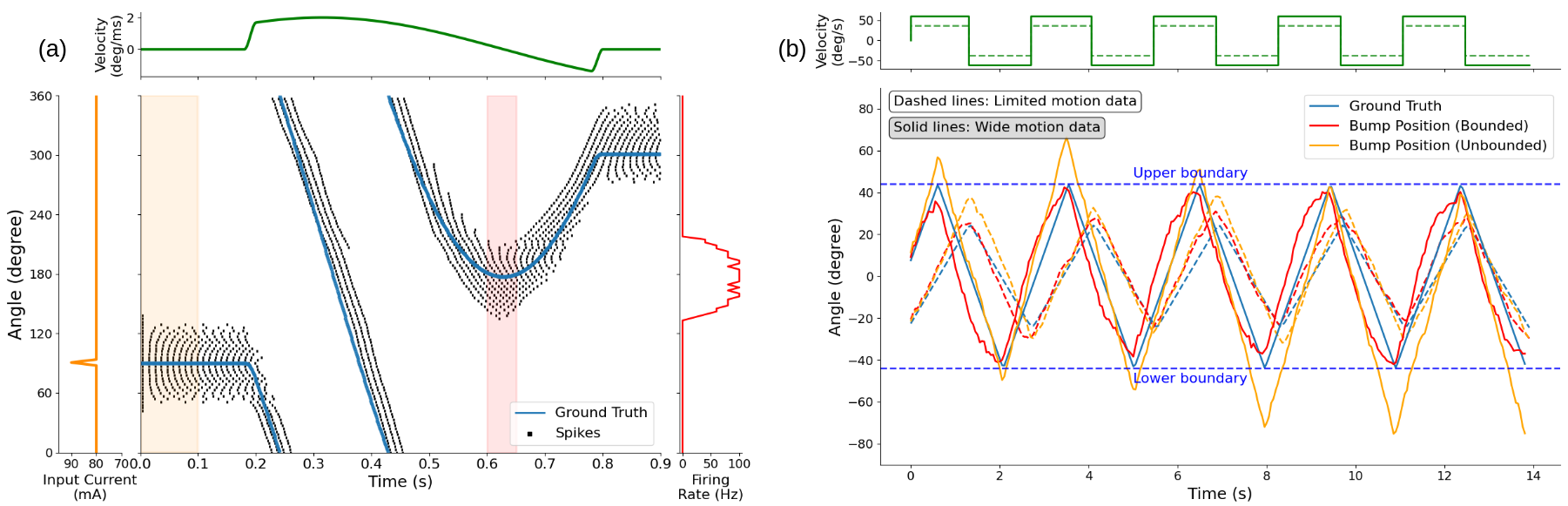}
\caption{\textbf{Neural encoding and validation of angular trajectory tracking.}
(a) Tracking using instantaneous velocity input. Raster plot of spiking activity (black) and ground–truth joint angle (blue). Top: input velocity; left: initialization current; right: mean firing–rate profile within the highlighted region.
(b) Tracking using target velocity commands only. Comparison between the unbounded (orange) and boundary–constrained (red) models for wide (solid) and limited (dashed) motion ranges. Incorporating boundary conditions reduces drift and improves alignment with ground truth.}
\label{fig:simulation}
\end{figure*}

\subsection{Network architecture}
The simulated network comprises \(N=120\) \ac{LIF} neurons~\cite{lapicqueQuantitativeInvestigationsElectrical2007}, each representing a preferred angle \(\theta_i\) over \([0, 2\pi)\).
All neurons receive a constant input current that maintains them just below firing threshold. At initialization, a brief additional current excites the neuron corresponding to the actual joint position, triggering local spiking and forming an activity bump that remains self-sustained through recurrent connectivity once the extra drive is removed.
Synaptic currents are exponentially decaying and driven by presynaptic spikes, weighted by two connectivity matrices; a symmetric term (\(W_{ij}^\mathrm{sym}\)) stabilizing the bump and a velocity-dependent antisymmetric term (\(W_{ij}^\mathrm{asym}\)) translating it along the ring:
\[
W^\mathrm{sym}_{ij}=g_\mathrm{inh}+g_\mathrm{cos}\cos(\theta_j-\theta_i),
W^\mathrm{asym}_{ij}=v(t)\,g_\mathrm{sin}\sin(\theta_j-\theta_i),
\]
where \(g_\mathrm{inh}=-16.46\) sets global inhibition, \(g_\mathrm{cos}=15.86\) controls bump width, and \(g_\mathrm{sin}=0.13\) is the optimized velocity-modulation gain (\figurename~\ref{fig:model_schematic}a--b).
The term \(v(t)\) scales both the sign and magnitude of the asymmetric coupling, shifting the bump proportionally to joint velocity.

To enforce mechanical limits, connectivity is further modulated near the joint’s physical range boundaries \(\theta_0\) and \(\theta_l\) (\figurename~\ref{fig:model_schematic}c), whose values depend on the specific joint being encoded.
Out-of-bound neurons (\(\theta_l<\theta_i<2\pi\)) form an unstable region centred at \(\theta_m^\ast = (\theta_0+\theta_l)/2+\pi\), preventing persistent activity beyond the valid range.
\[
W_{ij}^{\mathrm{asym}}=
\begin{cases}
v(t)\,g_{\sin}\,a^{+}({\theta_j})\,\sin(\theta_j-\theta_i), & v(t)>0\\[4pt]
v(t)\,g_{\sin}\,a^{-}({\theta_j})\,\sin(\theta_j-\theta_i), & v(t)<0
\end{cases}
\]
where \(a^{+}(\theta)\) and \(a^{-}(\theta)\) are piecewise linear attenuation functions that reshape the effective coupling near joint boundaries (\figurename~\ref{fig:model_schematic}d).
The joint angle estimate \(\hat{\theta}(t)\) is decoded in real time via a population vector average of the active neuron ensemble~\cite{georgopoulosNeuronalPopulationCoding1986}.

\subsection{Mapping the algorithm to DYNAP-SE}
We use the DYNAP-SE event-driven mixed-signal neuromorphic processor~\cite{Moradi_Qiao_Stefanini_Indiveri_2018}, with analog silicon neurons and synapses operating in biological real-time. The device has four cores, each with 256 adaptive-exponential integrate-and-fire neurons~\cite{Brette_Gerstner_2005, Chicca_Stefanini_Bartolozzi_Indiveri_2014}; synapses are configurable as fast/slow excitatory (AMPA/NMDA) or fast/slow inhibitory (GABA\_A/GABA\_B). Each neuron includes a 64-entry content-addressable memory for presynaptic addresses, and peripheral asynchronous digital logic handles address-event representation (AER) communication~\cite{Deiss_Douglas_Whatley_1998}. 
Resource constraints shaped the mapping. To fit the neuron budget and fan-in/fan-out limits, we used a discrete spiking ring attractor with ten populations of four neurons, providing adequate angular resolution within on-chip resources.
We implement cosine- and sine-shaped connectivity by varying the number of identical synapses rather than their individual strength; velocity inputs discretely add or remove direction-selective connections, translating the bump proportionally to $\lvert v \rvert$ (\figurename~\ref{fig:model_schematic}e–f). The DYNAP-SE's subthreshold analog dynamics natively emulate the underlying physical processes, allowing the neural architecture to intrinsically solve the system's temporal evolution.

\section{Results}
\subsection{Simulation results}

The ring–attractor network was implemented in the Brian simulator~\cite{Stimberg_Brette_Goodman_2019}, enabling efficient simulation of spiking dynamics with flexible connectivity.
We first tested whether the spiking ring–attractor without boundary conditions (\figurename~\ref{fig:model_schematic}a-b) could maintain a stable orientation estimate without continuous positional input. The network operated on a circular manifold representing a full $360^{\circ}$ joint with wrap-around continuity.
An input current encoding the initial joint position was applied for \(100\si{\milli\second}\), after which the network evolved autonomously, driven solely by velocity-dependent modulation of the asymmetric synapses, computed at each time-step from the motion profile.
As shown in~\figurename~\ref{fig:simulation}a, population activity rapidly formed a compact bump that persisted and tracked the true angular profile. The mean deviation between estimated angle and true joint angle remained below \(5^{\circ}\) over a \(900\si{\milli\second}\) simulation, confirming that the attractor preserves a consistent internal state without continuous input.
To test performance under realistic robotic conditions, the Brian model was coupled  with the MuJoCo physics engine~\cite{todorov2012mujoco} using the iCub humanoid model ~\cite{natale2017icub} to simulate the neck joint’s physical range of motion.
MuJoCo generated the ground-truth joint trajectories and the corresponding target velocity commands used to drive the simulation (\figurename~\ref{fig:simulation}b). Only target velocity signals were used, omitting acceleration and deceleration phases, forcing the attractor to reconstruct the trajectory from simplified control input. 
Simulations were run under two regimes: limited motion (within joint range) and wide motion (spanning full limits). Both network variants (~\figurename~\ref{fig:model_schematic}a-c ) tracked target trajectories accurately but differed in stability across conditions. 
In the limited-motion case, errors were low for both models (bounded: \(4.6^{\circ} \pm 3.2^{\circ}\); unbounded: \(5.2^{\circ} \pm 3.8^{\circ}\)), with no significant difference (\(p=0.14\)), though the bounded network showed a trend toward smaller drift and more stable error across 1s time windows. 
In the wide-motion case, boundary constraints clearly improved accuracy (\(10.7^{\circ} \pm 5.4^{\circ}\) vs.\ \(12.4^{\circ} \pm 8.6^{\circ}\); \(p=0.007\)), reducing cumulative error and variability near joint limits. Phase and time window analyses (1s bins) confirmed lower peak deviations and more consistent performance, indicating that boundary mechanisms enhance robustness even when only intermittently engaged.
Overall, the results show that the spiking ring–attractor sustains a stable internal representation of joint angle after a brief initialization period. Velocity inputs alone drive the bump dynamics along realistic trajectories, while boundary constraints enhance robustness and accuracy near joint limits.
Robustness was further assessed by varying network size and firing regime. Tracking accuracy saturated rapidly, reaching a stable mean absolute error of ~10° already for 60 neurons, consistent with the chosen 120-neuron configuration. Lower firing rates systematically degraded accuracy, while firing rates around 100 Hz minimized error.

\begin{figure*}[ht]
  \centering
 \includegraphics[width=1\textwidth,height=8cm]{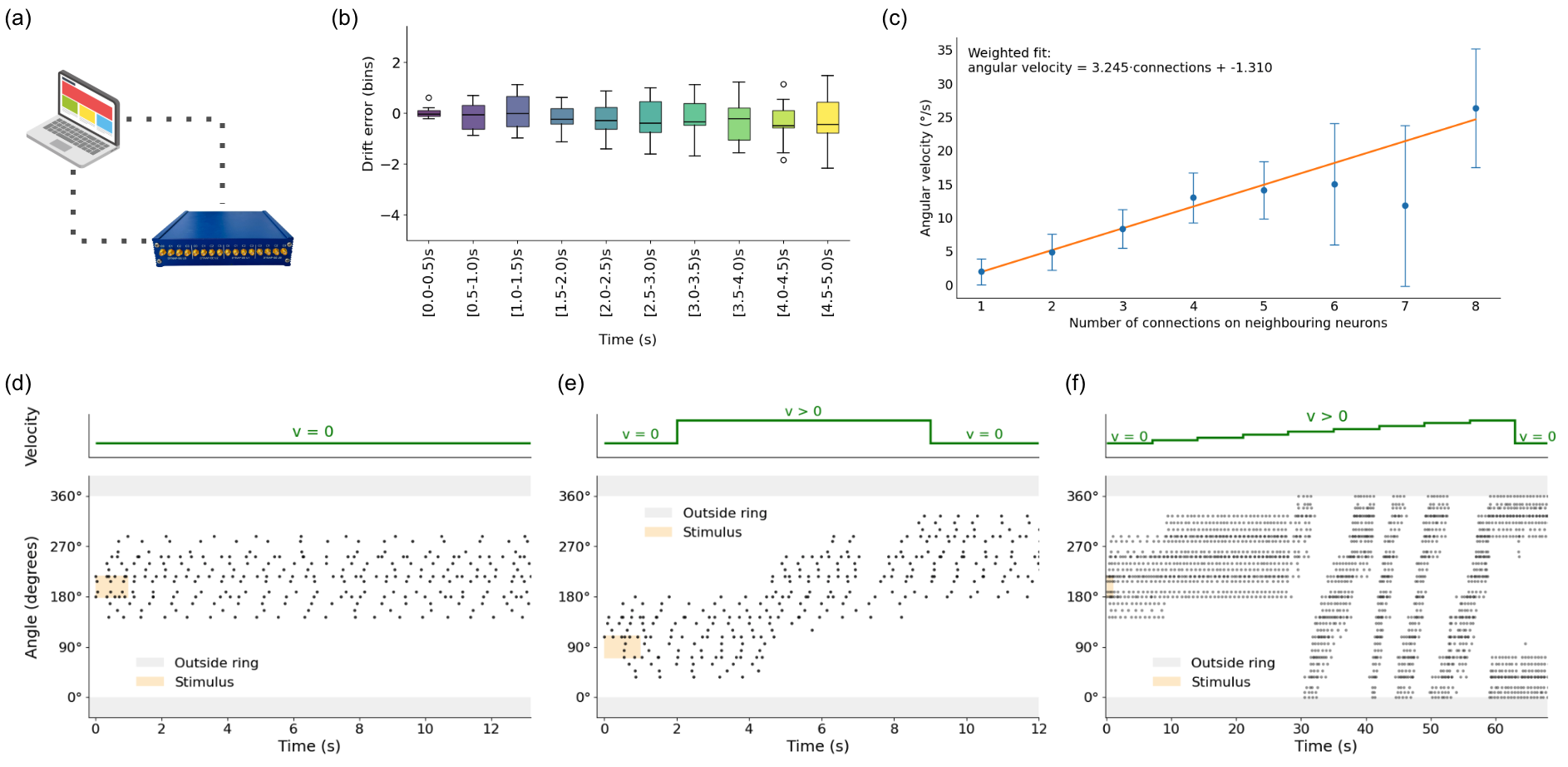}
 \caption{\textbf{Ring–attractor stability and velocity modulation on the DYNAP-SE neuromorphic processor.}
 (a) Experimental setup: a 1-\si{\second} Gaussian input is encoded by a LIF neuron and sent to the DYNAP-SE ring attractor; velocity connections are updated from the host computer.
 (b) Bump drift over time. Box plots show peak-position error across 0.5-s windows after stimulus offset (0–0.5 to 4.5–5.0 \si{\second}), pooled across ring populations; lower values indicate higher stability.
 (c) Mean bump angular velocity vs.\ number of velocity connections with weighted linear fit (\(w_i=1/\mathrm{SEM}_i^2\)). For each connection count, angular velocity was computed by unwrapping the bump trace and fitting a line over 2.5–9.0 \si{\second}.
 (d) Spike raster during 12\si{\second} baseline (\(v=0\)).
 (e) Spike raster during velocity modulation showing baseline (0-1\si{\second}), translation(2–9 \si{\second}) and recovery (9–12 \si{\second}).
 (f) Spike raster during “acceleration” test with stepwise addition and removal of velocity connections.}
  \label{fig:dynapse-prelim}
\end{figure*}

\subsection{Hardware validation results (DYNAP-SE)}
We tested whether a hardware ring attractor maintains a stable state estimate after an external cue and whether velocity inputs controllably translate the activity bump, consistent with simulation. The calibrated velocity-to-bump mapping is realized discretely by modulating the number of velocity connections (the chip-constrained analogue of the continuous gain). Drift stability and velocity–control linearity are interpreted as tests of the algorithm under mixed-signal noise, quantization, and routing limits. An example raster of stable bump activity is shown in~\figurename~\ref{fig:dynapse-prelim}d. To quantify drift, after a 1\si{\second} cue the bump was tracked for 5\si{\second}; drift error was the deviation of the peak firing position from the target population, summarized in 0.5\si{\second} windows and pooled across ring populations. Errors remained low across windows (Fig.~\ref{fig:dynapse-prelim}b), indicating a stable neck-angle representation without ongoing drive. To induce translation, velocity-modulated connections were introduced (\figurename~\ref{fig:dynapse-prelim}e,f). Three phases were tested: no velocity connections (0–2\si{\second}; stimulus 0–1\si{\second}), positive velocity (2–9\si{\second}), and removal of added velocity connections (9–12\si{\second}). The bump was stationary during baseline, advanced under positive velocity, and stabilised once the connections were removed (\figurename~\ref{fig:dynapse-prelim}e).
The dependence of velocity on the number of connections was then quantified. The number of velocity connections was varied measuring the resulting bump angular velocity. For each connection count, and a linear model was fit to the mean velocity over 10 repetition per population (\figurename~\ref{fig:dynapse-prelim}c). The mean velocity across populations and its \( 95\% \) confidence interval (mean \(\pm t\cdot\mathrm{SEM})\) were then computed. Finally, a linear model was fit across connection counts using inverse-variance weights \(w_i=1/\mathrm{SEM}_i^2\) (\figurename~\ref{fig:dynapse-prelim}d). Angular velocity increased approximately linearly with added connections.
Lastly, online adaptability was assessed by sequentially adding velocity connections and then removing them (``acceleration'' experiment;\figurename~\ref{fig:dynapse-prelim}f). As connections were added stepwise, the bump accelerated; after removal, the bump stabilised. Boundary constraints were evaluated in simulation and improved accuracy near limits; the present on-chip implementation omits boundary modulation and thus tests stability and velocity control only. 

\section{Conclusion}
This work demonstrates that compact spiking ring-attractor networks can encode robotic joint states in real-time and be deployed on neuromorphic hardware.
Driven only by target-velocity inputs, the unbounded network sustained a stable activity bump, while boundary modulation confined dynamics near joint limits and improved continuity through unmodelled motor transients.
On hardware, the network preserved post-cue stability and exhibited a linear relation between bump velocity and synaptic configuration, enabling predictable tuning of motion dynamics. The initialization current acts as an external sensory cue, analogous to how biological systems use multimodal input to recalibrate internal representations and limit path-integration drift, while the instantaneous velocity input parallels proprioceptive feedback from muscle spindles or joint sensors, anchoring bump dynamics to ongoing motion.
These results show that spiking attractors can achieve accurate proprioceptive estimation under realistic constraints.
Future work will investigate the system’s robustness in unstructured, noisy environments, testing the impact of stochastic synaptic noise, and sensory signal degradation. Incorporate boundary mechanisms, closed-loop sensory feedback and online calibration will extend the framework toward multi-joint, embodied neuromorphic control.

\section*{Acknowledgments}

This work began at the CapoCaccia 2025 Cognitive Neuromorphic Engineering Workshop. C.D.L. was supported by DSI, University of Zurich (G-95017-01-12). C.B. and F.M.F. were supported by the National Biodiversity Future Center (3138/2021, 3175/2021), PNRR MUR PE000013, and IIT Flagship “Brain and Machine Technologies for Sustainability.” B.M. and C.B. were supported by Horizon Europe PRIMI (101120727). A.M. was supported by MUR PNRR PE0000013–FAIR.
F.D. was supported by the Swiss National Science Foundation.

\bibliographystyle{IEEEtran}
\bibliography{localbib}

@article{Stimberg_Brette_Goodman_2019, title={Brian 2, an intuitive and efficient neural simulator}, volume={8}, ISSN={2050-084X}, DOI={10.7554/eLife.47314}, abstractNote={Brian 2 allows scientists to simply and efficiently simulate spiking neural network models. These models can feature novel dynamical equations, their interactions with the environment, and experimental protocols. To preserve high performance when defining new models, most simulators offer two options: low-level programming or description languages. The first option requires expertise, is prone to errors, and is problematic for reproducibility. The second option cannot describe all aspects of a computational experiment, such as the potentially complex logic of a stimulation protocol. Brian addresses these issues using runtime code generation. Scientists write code with simple and concise high-level descriptions, and Brian transforms them into efficient low-level code that can run interleaved with their code. We illustrate this with several challenging examples: a plastic model of the pyloric network, a closed-loop sensorimotor model, a programmatic exploration of a neuron model, and an auditory model with real-time input.}, journal={eLife}, author={Stimberg, Marcel and Brette, Romain and Goodman, Dan Fm}, year={2019}, month=aug, pages={e47314}, language={eng} }

@article{Moradi_Qiao_Stefanini_Indiveri_2018, title={A Scalable Multicore Architecture With Heterogeneous Memory Structures for Dynamic Neuromorphic Asynchronous Processors (DYNAPs)}, volume={12}, ISSN={1940-9990}, DOI={10.1109/TBCAS.2017.2759700}, abstractNote={Neuromorphic computing systems comprise networks of neurons that use asynchronous events for both computation and communication. This type of representation offers several advantages in terms of bandwidth and power consumption in neuromorphic electronic systems. However, managing the traffic of asynchronous events in large scale systems is a daunting task, both in terms of circuit complexity and memory requirements. Here, we present a novel routing methodology that employs both hierarchical and mesh routing strategies and combines heterogeneous memory structures for minimizing both memory requirements and latency, while maximizing programming flexibility to support a wide range of event-based neural network architectures, through parameter configuration. We validated the proposed scheme in a prototype multicore neuromorphic processor chip that employs hybrid analog/digital circuits for emulating synapse and neuron dynamics together with asynchronous digital circuits for managing the address-event traffic. We present a theoretical analysis of the proposed connectivity scheme, describe the methods and circuits used to implement such scheme, and characterize the prototype chip. Finally, we demonstrate the use of the neuromorphic processor with a convolutional neural network for the real-time classification of visual symbols being flashed to a dynamic vision sensor (DVS) at high speed.}, number={1}, journal={IEEE Transactions on Biomedical Circuits and Systems}, author={Moradi, Saber and Qiao, Ning and Stefanini, Fabio and Indiveri, Giacomo}, year={2018}, month=feb, pages={106–122} }

@article{Brette_Gerstner_2005, title={Adaptive exponential integrate-and-fire model as an effective description of neuronal activity}, volume={94}, ISSN={0022-3077}, DOI={10.1152/jn.00686.2005}, abstractNote={We introduce a two-dimensional integrate-and-fire model that combines an exponential spike mechanism with an adaptation equation, based on recent theoretical findings. We describe a systematic method to estimate its parameters with simple electrophysiological protocols (current-clamp injection of pulses and ramps) and apply it to a detailed conductance-based model of a regular spiking neuron. Our simple model predicts correctly the timing of 96% of the spikes (+/-2 ms) of the detailed model in response to injection of noisy synaptic conductances. The model is especially reliable in high-conductance states, typical of cortical activity in vivo, in which intrinsic conductances were found to have a reduced role in shaping spike trains. These results are promising because this simple model has enough expressive power to reproduce qualitatively several electrophysiological classes described in vitro.}, number={5}, journal={Journal of Neurophysiology}, author={Brette, Romain and Gerstner, Wulfram}, year={2005}, month=nov, pages={3637–3642}, language={eng} }

@article{Chicca_Stefanini_Bartolozzi_Indiveri_2014, title={Neuromorphic Electronic Circuits for Building Autonomous Cognitive Systems}, volume={102}, ISSN={1558-2256}, DOI={10.1109/JPROC.2014.2313954}, abstractNote={Several analog and digital brain-inspired electronic systems have been recently proposed as dedicated solutions for fast simulations of spiking neural networks. While these architectures are useful for exploring the computational properties of large-scale models of the nervous system, the challenge of building low-power compact physical artifacts that can behave intelligently in the real world and exhibit cognitive abilities still remains open. In this paper, we propose a set of neuromorphic engineering solutions to address this challenge. In particular, we review neuromorphic circuits for emulating neural and synaptic dynamics in real time and discuss the role of biophysically realistic temporal dynamics in hardware neural processing architectures; we review the challenges of realizing spike-based plasticity mechanisms in real physical systems and present examples of analog electronic circuits that implement them;we describe the computational properties of recurrent neural networks and show how neuromorphic winner-take-all circuits can implement working-memory and decision-making mechanisms. We validate the neuromorphic approach proposed with experimental results obtained from our own circuits and systems, and argue how the circuits and networks presented in this work represent a useful set of components for efficiently and elegantly implementing neuromorphic cognition.}, number={9}, journal={Proceedings of the IEEE}, author={Chicca, Elisabetta and Stefanini, Fabio and Bartolozzi, Chiara and Indiveri, Giacomo}, year={2014}, month=sep, pages={1367–1388} }

@inbook{Deiss_Douglas_Whatley_1998, title={A Pulse-Coded Communications Infrastructure for Neuromorphic Systems}, ISBN={978-0-262-27876-8}, url={https://direct.mit.edu/books/book/2001/chapter/54534/A-Pulse-Coded-Communications-Infrastructure-for}, DOI={10.7551/mitpress/5704.003.0011}, booktitle={Pulsed Neural Networks}, publisher={The MIT Press}, author={Deiss, Stephen R. and Douglas, Rodney J. and Whatley, Adrian M.}, editor={Maass, Wolfgang and Bishop, Christopher M.}, year={1998}, month=nov, pages={157–178}, language={en} }

@article{noormanMaintainingUpdatingAccurate2024,
  title = {Maintaining and Updating Accurate Internal Representations of Continuous Variables with a Handful of Neurons},
  author = {Noorman, Marcella and Hulse, Brad K. and Jayaraman, Vivek and Romani, Sandro and Hermundstad, Ann M.},
  year = {2024},
  month = nov,
  journal = {Nature Neuroscience},
  volume = {27},
  number = {11},
  pages = {2207--2217},
  publisher = {Nature Publishing Group},
  issn = {1546-1726},
  doi = {10.1038/s41593-024-01766-5},
  urldate = {2025-03-11},
  copyright = {2024 The Author(s)},
  langid = {english}
}

@article{kreiser2020chip,
  title={An on-chip spiking neural network for estimation of the head pose of the icub robot},
  author={Kreiser, Raphaela and Renner, Alpha and Leite, Vanessa RC and Serhan, Baris and Bartolozzi, Chiara and Glover, Arren and Sandamirskaya, Yulia},
  journal={Frontiers in Neuroscience},
  volume={14},
  pages={551},
  year={2020},
  publisher={Frontiers Media SA}
}

@article{stone2017anatomically,
  title={An anatomically constrained model for path integration in the bee brain},
  author={Stone, Thomas and Webb, Barbara and Adden, Andrea and Weddig, Nicolai Ben and Honkanen, Anna and Templin, Rachel and Wcislo, William and Scimeca, Luca and Warrant, Eric and Heinze, Stanley},
  journal={Current Biology},
  volume={27},
  number={20},
  pages={3069--3085},
  year={2017},
  publisher={Elsevier}
}

@article{olofsson2015sensor,
  title={Sensor fusion for robotic workspace state estimation},
  author={Olofsson, Bj{\"o}rn and Antonsson, Jacob and Kortier, Henk G and Bernhardsson, Bo and Robertsson, Anders and Johansson, Rolf},
  journal={IEEE/ASME Transactions on Mechatronics},
  volume={21},
  number={5},
  pages={2236--2248},
  year={2015},
  publisher={IEEE}
}

@article{bartolozzi2022embodied,
  title={Embodied neuromorphic intelligence},
  author={Bartolozzi, Chiara and Indiveri, Giacomo and Donati, Elisa},
  journal={Nature communications},
  volume={13},
  number={1},
  pages={1024},
  year={2022},
  publisher={Nature Publishing Group UK London}
}

@article{chicca2014neuromorphic,
  title={Neuromorphic electronic circuits for building autonomous cognitive systems},
  author={Chicca, Elisabetta and Stefanini, Fabio and Bartolozzi, Chiara and Indiveri, Giacomo},
  journal={Proceedings of the IEEE},
  volume={102},
  number={9},
  pages={1367--1388},
  year={2014},
  publisher={IEEE}
}

@article{seelig2015neural,
  title={Neural dynamics for landmark orientation and angular path integration},
  author={Seelig, Johannes D and Jayaraman, Vivek},
  journal={Nature},
  volume={521},
  number={7551},
  pages={186--191},
  year={2015},
  publisher={Nature Publishing Group UK London}
}

@article{fisher2019sensorimotor,
  title={Sensorimotor experience remaps visual input to a heading-direction network},
  author={Fisher, Yvette E and Lu, Jenny and D’Alessandro, Isabel and Wilson, Rachel I},
  journal={Nature},
  volume={576},
  number={7785},
  pages={121--125},
  year={2019},
  publisher={Nature Publishing Group UK London}
}

@inproceedings{todorov2012mujoco,
  title={MuJoCo: A physics engine for model-based control},
  author={Todorov, Emanuel and Erez, Tom and Tassa, Yuval},
  booktitle={2012 IEEE/RSJ International Conference on Intelligent Robots and Systems},
  pages={5026--5033},
  year={2012},
  organization={IEEE},
  doi={10.1109/IROS.2012.6386109}
}

@article{Song_Wang_2005, title={Angular Path Integration by Moving “Hill of Activity”: A Spiking Neuron Model without Recurrent Excitation of the Head-Direction System}, volume={25}, rights={Copyright © 2005 Society for Neuroscience 0270-6474/05/251002-13.00/0}, ISSN={0270-6474, 1529-2401}, DOI={10.1523/JNEUROSCI.4172-04.2005}, abstractNote={During spatial navigation, the head orientation of an animal is encoded internally by neural persistent activity in the head-direction (HD) system. In computational models, such a bell-shaped “hill of activity” is commonly assumed to be generated by recurrent excitation in a continuous attractor network. Recent experimental evidence, however, indicates that HD signal in rodents originates in a reciprocal loop between the lateral mammillary nucleus (LMN) and the dorsal tegmental nucleus (DTN), which is characterized by a paucity of local excitatory axonal collaterals. Moreover, when the animal turns its head to a new direction, the heading information is updated by a time integration of angular head velocity (AHV) signals; the underlying mechanism remains unresolved. To investigate these issues, we built and investigated an LMN-DTN network model that consists of three populations of noisy and spiking neurons coupled by biophysically realistic synapses. We found that a combination of uniform external excitation and recurrent cross-inhibition can give rise to direction-selective persistent activity. The model reproduces the experimentally observed three types of HD tuning curves differentially modulated by AHV and anticipatory firing activity in LMN HD cells. Time integration is assessed by using constant or sinusoidal angular velocity stimuli, as well as naturalistic AHV inputs (from rodent recordings). Furthermore, the internal representation of head direction is shown to be calibrated or reset by strong external cues. We identify microcircuit properties that determine the ability of our model network to subserve time integration function.}, number={4}, journal={Journal of Neuroscience}, publisher={Society for Neuroscience}, author={Song, Pengcheng and Wang, Xiao-Jing}, year={2005}, month=jan, pages={1002–1014}, language={en} }

@article{Seeholzer_Deger_Gerstner_2019, title={Stability of working memory in continuous attractor networks under the control of short-term plasticity}, volume={15}, ISSN={1553-7358}, DOI={10.1371/journal.pcbi.1006928}, abstractNote={Continuous attractor models of working-memory store continuous-valued information in continuous state-spaces, but are sensitive to noise processes that degrade memory retention. Short-term synaptic plasticity of recurrent synapses has previously been shown to affect continuous attractor systems: short-term facilitation can stabilize memory retention, while short-term depression possibly increases continuous attractor volatility. Here, we present a comprehensive description of the combined effect of both short-term facilitation and depression on noise-induced memory degradation in one-dimensional continuous attractor models. Our theoretical description, applicable to rate models as well as spiking networks close to a stationary state, accurately describes the slow dynamics of stored memory positions as a combination of two processes: (i) diffusion due to variability caused by spikes; and (ii) drift due to random connectivity and neuronal heterogeneity. We find that facilitation decreases both diffusion and directed drifts, while short-term depression tends to increase both. Using mutual information, we evaluate the combined impact of short-term facilitation and depression on the ability of networks to retain stable working memory. Finally, our theory predicts the sensitivity of continuous working memory to distractor inputs and provides conditions for stability of memory.}, number={4}, journal={PLOS Computational Biology}, publisher={Public Library of Science}, author={Seeholzer, Alexander and Deger, Moritz and Gerstner, Wulfram}, year={2019}, month=apr, pages={e1006928}, language={en} }

@article{Kim_Rouault_Druckmann_Jayaraman_2017, title={Ring attractor dynamics in the Drosophila central brain}, volume={356}, ISSN={1095-9203}, DOI={10.1126/science.aal4835}, abstractNote={Ring attractors are a class of recurrent networks hypothesized to underlie the representation of heading direction. Such network structures, schematized as a ring of neurons whose connectivity depends on their heading preferences, can sustain a bump-like activity pattern whose location can be updated by continuous shifts along either turn direction. We recently reported that a population of fly neurons represents the animal’s heading via bump-like activity dynamics. We combined two-photon calcium imaging in head-fixed flying flies with optogenetics to overwrite the existing population representation with an artificial one, which was then maintained by the circuit with naturalistic dynamics. A network with local excitation and global inhibition enforces this unique and persistent heading representation. Ring attractor networks have long been invoked in theoretical work; our study provides physiological evidence of their existence and functional architecture.}, number={6340}, journal={Science (New York, N.Y.)}, author={Kim, Sung Soo and Rouault, Hervé and Druckmann, Shaul and Jayaraman, Vivek}, year={2017}, month=may, pages={849–853}, language={eng} }

@inproceedings{Tang_Shah_Michmizos_2019, address={Macau, China}, title={Spiking Neural Network on Neuromorphic Hardware for Energy-Efficient Unidimensional SLAM}, url={https://doi.org/10.1109/IROS40897.2019.8967864}, DOI={10.1109/IROS40897.2019.8967864}, abstractNote={Energy-efficient simultaneous localization and mapping (SLAM) is crucial for mobile robots exploring unknown environments. The mammalian brain solves SLAM via a network of specialized neurons, exhibiting asynchronous computations and event-based communications, with very low energy consumption. We propose a brain-inspired spiking neural network (SNN) architecture that solves the unidimensional SLAM by introducing spike-based reference frame transformation, visual likelihood computation, and Bayesian inference. We integrated our neuromorphic algorithm to Intel&amp;#x2019;s Loihi neuromorphic processor, a non-Von Neumann hardware that mimics the brain&amp;#x2019;s computing paradigms. We performed comparative analyses for accuracy and energy-efficiency between our neuromorphic approach and the GMapping algorithm, which is widely used in small environments. Our Loihi-based SNN architecture consumes 100 times less energy than GMapping run on a CPU while having comparable accuracy in head direction localization and map-generation. These results pave the way for scaling our approach towards active-SLAM alternative solutions for Loihi-controlled autonomous robots.}, booktitle={2019 IEEE/RSJ International Conference on Intelligent Robots and Systems (IROS)}, publisher={IEEE Press}, author={Tang, Guangzhi and Shah, Arpit and Michmizos, Konstantinos P.}, year={2019}, month=nov, pages={4176–4181} }

@article{Zhao_Risi_Monforte_Bartolozzi_Indiveri_Donati_2020, title={Closed-Loop Spiking Control on a Neuromorphic Processor Implemented on the iCub}, volume={10}, ISSN={2156-3365}, DOI={10.1109/JETCAS.2020.3040390}, abstractNote={Neuromorphic engineering promises the deployment of low latency, adaptive and low power systems that can lead to the design of truly autonomous artificial agents. However, many building blocks for developing a fully neuromorphic artificial agent are still missing. While neuromorphic sensing, perception, and decision-making building blocks are quite mature, the ones for motor control and actuation are lagging behind. In this paper we present a closed-loop motor controller implemented on a mixed-signal analog/digital neuromorphic processor which emulates a spiking neural network that continuously calculates an error signal from the desired target and the feedback signals. The system uses population coding and recurrent Winner-Take-All networks to encode the signals robustly. Recurrent connections within each population are used to speed up the convergence, decrease the effect of mismatch and improve selectivity. The error signal computed in this way is then fed into three additional populations of spiking neurons which produce the proportional, integral and derivative terms of classical controllers exploiting the temporal dynamics of the network synapses and neurons. To validate this approach we interfaced this neuromorphic motor controller with an iCub robot simulator. We tested our spiking controller in a single joint control task for the robot head yaw. We demonstrate the correct performance of the spiking controller in a step response experiment and apply it to a target pursuit task.}, number={4}, journal={IEEE Journal on Emerging and Selected Topics in Circuits and Systems}, author={Zhao, Jingyue and Risi, Nicoletta and Monforte, Marco and Bartolozzi, Chiara and Indiveri, Giacomo and Donati, Elisa}, year={2020}, month=dec, pages={546–556} }

@article{Richter_Wu_Whatley_Köstinger_Nielsen_Qiao_Indiveri_2024, title={DYNAP-SE2: a scalable multi-core dynamic neuromorphic asynchronous spiking neural network processor}, volume={4}, ISSN={2634-4386}, DOI={10.1088/2634-4386/ad1cd7}, abstractNote={With the remarkable progress that technology has made, the need for processing data near the sensors at the edge has increased dramatically. The electronic systems used in these applications must process data continuously, in real-time, and extract relevant information using the smallest possible energy budgets. A promising approach for implementing always-on processing of sensory signals that supports on-demand, sparse, and edge-computing is to take inspiration from biological nervous system. Following this approach, we present a brain-inspired platform for prototyping real-time event-based spiking neural networks. The system proposed supports the direct emulation of dynamic and realistic neural processing phenomena such as short-term plasticity, NMDA gating, AMPA diffusion, homeostasis, spike frequency adaptation, conductance-based dendritic compartments and spike transmission delays. The analog circuits that implement such primitives are paired with a low latency asynchronous digital circuits for routing and mapping events. This asynchronous infrastructure enables the definition of different network architectures, and provides direct event-based interfaces to convert and encode data from event-based and continuous-signal sensors. Here we describe the overall system architecture, we characterize the mixed signal analog-digital circuits that emulate neural dynamics, demonstrate their features with experimental measurements, and present a low- and high-level software ecosystem that can be used for configuring the system. The flexibility to emulate different biologically plausible neural networks, and the chip’s ability to monitor both population and single neuron signals in real-time, allow to develop and validate complex models of neural processing for both basic research and edge-computing applications.}, number={1}, journal={Neuromorphic Computing and Engineering}, publisher={IOP Publishing}, author={Richter, Ole and Wu, Chenxi and Whatley, Adrian M and Köstinger, German and Nielsen, Carsten and Qiao, Ning and Indiveri, Giacomo}, year={2024}, month=jan, pages={014003}, language={en} }

@article{Samsonovich_McNaughton_1997, title={Path Integration and Cognitive Mapping in a Continuous Attractor Neural Network Model}, volume={17}, ISSN={0270-6474}, DOI={10.1523/JNEUROSCI.17-15-05900.1997}, abstractNote={A minimal synaptic architecture is proposed for how the brain might perform path integration by computing the next internal representation of self-location from the current representation and from the perceived velocity of motion. In the model, a place-cell assembly called a “chart” contains a two-dimensional attractor set called an “attractor map” that can be used to represent coordinates in any arbitrary environment, once associative binding has occurred between chart locations and sensory inputs. In hippocampus, there are different spatial relations among place fields in different environments and behavioral contexts. Thus, the same units may participate in many charts, and it is shown that the number of uncorrelated charts that can be encoded in the same recurrent network is potentially quite large. According to this theory, the firing of a given place cell is primarily a cooperative effect of the activity of its neighbors on the currently active chart. Therefore, it is not particularly useful to think of place cells as encoding any particular external object or event. Because of its recurrent connections, hippocampal field CA3 is proposed as a possible location for this “multichart” architecture; however, other implementations in anatomy would not invalidate the main concepts. The model is implemented numerically both as a network of integrate-and-fire units and as a “macroscopic” (with respect to the space of states) description of the system, based on a continuous approximation defined by a system of stochastic differential equations. It provides an explanation for a number of hitherto perplexing observations on hippocampal place fields, including doubling, vanishing, reshaping in distorted environments, acquiring directionality in a two-goal shuttling task, rapid formation in a novel environment, and slow rotation after disorientation. The model makes several new predictions about the expected properties of hippocampal place cells and other cells of the proposed network.}, number={15}, journal={The Journal of Neuroscience}, author={Samsonovich, Alexei and McNaughton, Bruce L.}, year={1997}, month=aug, pages={5900–5920} }

@article{lapicqueQuantitativeInvestigationsElectrical2007,
  title = {Quantitative Investigations of Electrical Nerve Excitation Treated as Polarization. 1907},
  author = {Lapicque, Louis},
  year = {2007},
  month = dec,
  journal = {Biological Cybernetics},
  volume = {97},
  number = {5-6},
  pages = {341--349},
  issn = {0340-1200},
  doi = {10.1007/s00422-007-0189-6},
  langid = {english},
  pmid = {18046573}
}

@article{falotico2017head,
  title={Head stabilization in a humanoid robot: models and implementations},
  author={Falotico, Egidio and Cauli, Nino and Kryczka, Przemyslaw and Hashimoto, Kenji and Berthoz, Alain and Takanishi, Atsuo and Dario, Paolo and Laschi, Cecilia},
  journal={Autonomous Robots},
  volume={41},
  number={2},
  pages={349--365},
  year={2017},
  publisher={Springer}
}

@article{georgopoulosNeuronalPopulationCoding1986,
  title = {Neuronal {{Population Coding}} of {{Movement Direction}}},
  author = {Georgopoulos, Apostolos P. and Schwartz, Andrew B. and Kettner, Ronald E.},
  year = {1986},
  month = sep,
  journal = {Science},
  volume = {233},
  number = {4771},
  pages = {1416--1419},
  publisher = {American Association for the Advancement of Science},
  doi = {10.1126/science.3749885},
  urldate = {2025-10-10}
}

@article{natale2017icub,
  title={iCub: The not-yet-finished story of building a robot child},
  author={Natale, Lorenzo and Bartolozzi, Chiara and Pucci, Daniele and Wykowska, Agnieszka and Metta, Giorgio},
  journal={Science Robotics},
  volume={2},
  number={13},
  pages={eaaq1026},
  year={2017},
  publisher={American Association for the Advancement of Science}
}

\end{document}